\documentclass[runningheads]{llncs}
\usepackage{amsmath}
\usepackage{amssymb}
\usepackage{graphicx}
\usepackage{comment}
\usepackage{makecell}
\usepackage{caption}
\usepackage{multirow}
\usepackage{subcaption}
\usepackage{xcolor}
\usepackage{sidecap}
\usepackage{wrapfig}
\usepackage{svg}

\usepackage[ruled,noend]{algorithm2e}
\usepackage[noend]{algpseudocode}
\SetKwInput{Input}{Input}                
\SetKwInput{Output}{Output\,}              

\SetCommentSty{mycommfont}

\usepackage{mathtools}
\DeclarePairedDelimiter\floor{\lfloor}{\rfloor}

\begin{document}
\title{Obtaining Smoothly Navigable Approximation Sets in Bi-Objective Multi-Modal Optimization}
\titlerunning{Bi-Objective Multi-Modal Bézier curve parameterizations.}
%
%
\author{Renzo J. Scholman\inst{1,3} \orcidID{0000-0003-2813-015X} \and
Anton Bouter\inst{1} \orcidID{0000-0003-4599-0733} \and
Leah R.M. Dickhoff\inst{2} \orcidID{0000-0001-6720-4380} \thanks{Leah R.M. Dickhoff was supported by the Dutch Cancer Society (KWF Kankerbestrijding, Project N.12183) and Elekta.} \and \\
Tanja Alderliesten\inst{2} \orcidID{0000-0003-4261-7511} \and \\
Peter A.N. Bosman\inst{1,3} \orcidID{0000-0002-4186-6666}}
\authorrunning{R. Scholman et al.}
%
\institute{Centrum Wiskunde \& Informatica, Amsterdam, The Netherlands \\
\email{\{Renzo.Scholman,Anton.Bouter,Peter.Bosman\}@cwi.nl}
\and
Leiden University Medical Center, Leiden, The Netherlands \\
\email{\{L.R.M.Dickhoff,T.Alderliesten\}@lumc.nl} \and
Delft University of Technology, Delft, The Netherlands}
\maketitle              
\begin{abstract}
Even if a Multi-modal Multi-Objective Evolutionary Algorithm (MMOEA) is designed to find solutions well spread over all locally optimal approximation sets of a Multi-modal Multi-objective Optimization Problem (MMOP), there is a risk that the found set of solutions is not smoothly navigable because the solutions belong to various niches, reducing the insight for decision makers.
To tackle this issue, a new MMOEAs is proposed: the Multi-Modal Bézier Evolutionary Algorithm (MM-BezEA), which produces approximation sets that cover individual niches and exhibit inherent decision-space smoothness as they are parameterized by Bézier curves.
MM-BezEA combines the concepts behind the recently introduced BezEA and MO-HillVallEA to find all locally optimal approximation sets.
When benchmarked against the MMOEAs MO\_Ring\_PSO\_SCD and MO-HillVallEA on MMOPs with linear Pareto sets, MM-BezEA was found to perform best in terms of best hypervolume.  

\end{abstract}

\section{Introduction}
Many real-world optimization problems have multiple conflicting objectives, whereby improvement in one objective often results in the deterioration of another.
Multi-Objective Evolutionary Algorithms (MOEAs), like NSGA-II \cite{nsga2}, MOEA/D \cite{igdx}, and MO-CMA-ES \cite{MO-CMA-ES}, are widely accepted to be well-suited to solve such Multi-objective Optimization Problems (MOPs) \cite{deb_moea}. 
The aim is to obtain a set of solutions, called the approximation set, such that all solutions are non-dominated and the set itself is close to the set of Pareto-optimal solutions.
Here, a solution $\boldsymbol{x}_0$ dominates $\boldsymbol{x}_1$ ($\boldsymbol{x}_0 \succ \boldsymbol{x}_1$) in an MOP with $m$ objectives if $\forall i \in \{0,1,...,m-1\}: f_i(\boldsymbol{x}_0) \leq f_i(\boldsymbol{x}_1)$ and $\exists i \in \{0,1,...,m-1\}: f_i(\boldsymbol{x}_0) < f_i(\boldsymbol{x}_1)$.
The Pareto Set (PS) is $\mathcal{P}_S = \{\boldsymbol{x}_i | \neg \exists \boldsymbol{x}_j: \boldsymbol{x}_j \succ \boldsymbol{x}_i\}$ and the Pareto Front (PF) is $\mathcal{P}_F = \left\{\left(f_0(\boldsymbol{x}), \cdots, f_{m-1}(\boldsymbol{x})\right) | \boldsymbol{x} \in \mathcal{P}_S\right\}$.

A more complex type of MOPs is that of Multi-modal MOPs (MMOPs), where the goal is not to find one, but multiple, if not all, (local) PSs.
In MMOPs, each of the PSs pertains to a \textit{niche} , a subset of the search space, where a single mode resides, i.e., with one local PS.
The PSs may, however, map to the same PF in objective space, similar to having multiple (locally) optimal solutions of the same quality in a single-objective problem, e.g., the sine function.
Here we consider MMOPs in the case of real-valued parameters, or continuous optimization.
This field has recently gotten more traction, with reviews \cite{tanabe_review}, proposed formal definitions \cite{mmop} and new visualization techniques \cite{mmop_viz}. 

In order to have MOEAs solve MMOPs, they need additional tools that prevent their convergence to a single niche in the landscape \cite{Mahfoud95nichingmethods}.
Niching \cite{niching} is one of such diversity-preserving tools used by Multi-modal MOEAs (MMOEAs) to effectively and simultaneously search for solutions near the (local) PS in each niche.
Niching has been successfully applied to established MMOEAs in the form of the multi-objective particle swarm optimization using ring topology and special crowding distance (MO\_Ring\_PSO\_SCD) algorithm \cite{mo-pso-ring} and Omni-optimizer \cite{omni-optimizer} among others.

\vspace{-5mm}
\begin{figure}
  \setlength\abovecaptionskip{0.2\baselineskip}
  \setlength{\belowcaptionskip}{-10pt}
  \centering
  \includegraphics[width=\linewidth]{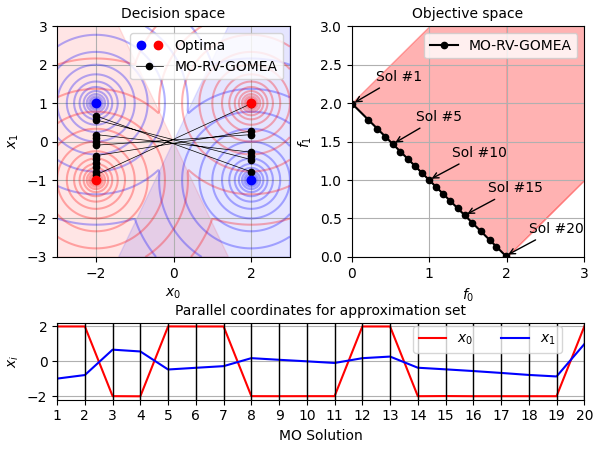}
  \caption{Approximation set and front with parallel coordinates plot as produced by MO-RV-GOMEA on the MinDist problem: $f_0(\boldsymbol{x}) = min(||\boldsymbol{x} - [1,-1]||, ||\boldsymbol{x} - [-1,1]||)$ and $f_1(\boldsymbol{x}) = min(||\boldsymbol{x} - [1,1]||, ||\boldsymbol{x} - [-1,-1]||)$. Shaded blue and red regions correspond to niches with global PSs.}
  \label{fig:problem}
\end{figure}

Most MMOEAs do not explicitly model multiple approximation sets, but include diversity preserving techniques to ensure that solutions from multiple niches are maintained.
The result of these MMOEAs is usually given in the form of a single approximation front, often derived from (a subset of) the elitist archive.
A decision maker can then investigate this front by traversing the solutions for desired trade-offs.
However, the underlying solutions are taken from several distinct niches, which could result in observing a counterintuitive change in decision variable values when navigating the approximation front.
Decision makers then might have to investigate all solutions before a correct choice can be made \cite{Luong2018,bezea}.
Figure \ref{fig:problem} shows such an approximation front and set that contains solutions from both modes on the MinDist problem \cite{mohillvallea} and demonstrates the counterintuitive changes in decision variable values in the parallel coordinates plot (i.e., if one were to navigate the front by traversing and inspecting the solutions from one extreme to the other).
It shows that in objective space a front is found that looks to have approximated the PF to (near) optimality, but the solutions jump around throughout decision space as seen in the parallel coordinates plot.

The issue of counterintuitive navigation along the approximation front has also been explored in recent work, which introduced a new indicator-based MOEA for bi-objective optimization called BezEA \cite{bezea}. 
A new problem formulation for population-based MOEAs was introduced whereby they parameterized approximation sets as Bézier curves.
This formulation ensures the navigational smoothness of an approximation set, whilst still being able to find good approximation sets when using the HyperVolume indicator (HV) \cite{spea2}.
By design, BezEA disallowes curves to dominate parts of themselves to ensure that the approximation set constitutes a single niche in the landscape.

Recent work that included the concept of niching showed promising results in maintaining multiple approximation sets in a popula\-tion-based MMOEA called the MO-HillVallEA \cite{mohillvallea}.
The authors extended the concept of Hill-Valley Clustering (HVC) \cite{hillvallea} for MOPs to Multi-Objective HVC (MO-HVC) for MMOPs.
MO-HillVallEA was found to be capable of finding and preserving approximation sets, one for each niche, in parallel over time by considering Pareto domination per niche.
However, MO-HillVallEA produces approximation sets that are not inherently smooth due to slight oscillations around the PS. 

In this work, the notions of niching through HVC and Bézier curve parameterizations are combined. 
The use of niching allows to effectively search the multi-modal landscape.
The use of Bézier curve parameterizations not only enforces the smooth and intuitive navigability that is desired by decision makers, but also enforces each approximation set to be within a single niche.
Furthermore, the use of the HV indicator allows closer convergence to the PS as compared to the Pareto dominance-based algorithms \cite{hv_vs_pareto}.
The new algorithm that we propose is called Multi-Modal Bézier Evolutionary Algorithm (MM-BezEA).
The purpose of MM-BezEA is to find all approximation sets for a given MMOP, where each approximation set consists of solutions from a single mode.

In order to combine the techniques of Bézier curve parameterizations and HVC into the proposed algorithm, several contributions are made.
First, the problem of how to niche approximation sets in the form of Bézier curve parameterizations is resolved.
Second, initialization of approximation sets within a single niche is enabled, as otherwise, clustering becomes ambiguous if these approximation sets span multiple niches.

\section{Bézier parameterizations}
One of the key features of the newly proposed algorithm is that Bézier parameterizations are used as approximation sets for bi-objective optimization \cite{bezea}.
This allows the algorithm to model the approximation set as a smooth curve in decision space.

\subsection{Definition of solution set}
\begin{figure}
  \setlength\abovecaptionskip{0.2\baselineskip}
  \centering
  \includegraphics[width=0.6\linewidth]{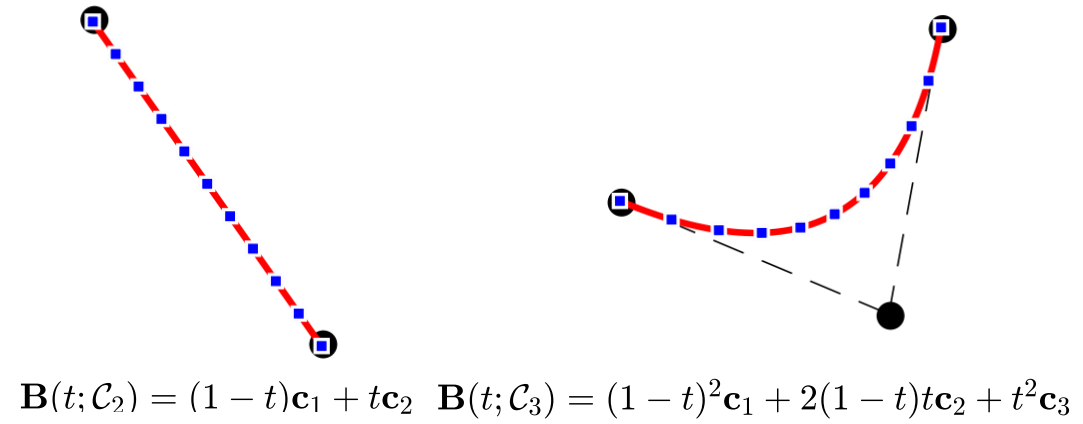}
  \caption{Bézier curves with $q \in \{2,3\}$ control points in black. Interpolated curve in red with the $p = 11$ points in blue evenly spread in the domain of $t$ along the curve \cite{bezea}.}
  \label{fig:bezier}
\end{figure}

An $\ell$-dimensional Bézier curve $\mathbf{B}(t;C_q)$ can be defined using $q \geq 2$ control points $c_i$ in an ordered set $C_q = \left\{\boldsymbol{c}_1,...,\boldsymbol{c}_q\right\}$, where $\ell$ is the problem dimensionality and $\boldsymbol{c}_i \in \mathbb{R}^\ell$. 
The full notation is:
\begin{equation}
\small
    \mathbf{B}(t;C_q) = \sum_{i=1}^q \binom{q - 1}{i - 1} (1-t)^{q-i-1}t^{i-1} \boldsymbol{c}_i \text{ for } 0 \leq t \leq 1
\end{equation}
The endpoints of the Bézier curve are always defined by the first and last control points, whilst the other control points are normally not located on the curve.
A solution set of given size $p$, $S_{p,q}(C_q) = \left\{\boldsymbol{x}_1,...,\boldsymbol{x}_p\right\}$ with $\boldsymbol{x}_i \in \mathbb{R}^\ell$, can now be parameterized by a Bézier curve by selecting an evenly spread set of $p$ points $\boldsymbol{x}_i$.
Figure \ref{fig:bezier} visualizes two solution sets $S_{p,q}(C_q)$ parameterized by Bézier curves.
The solution set $S_{p,q}(C_q)$ is formally defined as:
\begin{equation}
\small
    S_{p,q}\left(C_q\right) = \left\{ \mathbf{B}\left(\dfrac{0}{p-1};C_q\right), \mathbf{B}\left(\dfrac{1}{p-1};C_q\right), ..., \mathbf{B}\left(\dfrac{p-1}{p-1};C_q\right) \right\}
    \label{eq:sol_set}
\end{equation}
$S_{p,q}(C_q)$ is parameterized for (M)MOEAs by taking the concatenation of the decision variables in the set of control points $C_q$ as a solution \cite{bezea,smsemoa}.
This results in a solution being of the form $[\boldsymbol{c}_1,...,\boldsymbol{c}_q] \in \mathbb{R}^{q \times \ell}$.

\vspace{-9mm}

\subsection{Evaluation}
\begin{wrapfigure}{r}{0.65\textwidth}
  \setlength{\belowcaptionskip}{-10pt}
  \centering
  \includegraphics[width=1.0\linewidth]{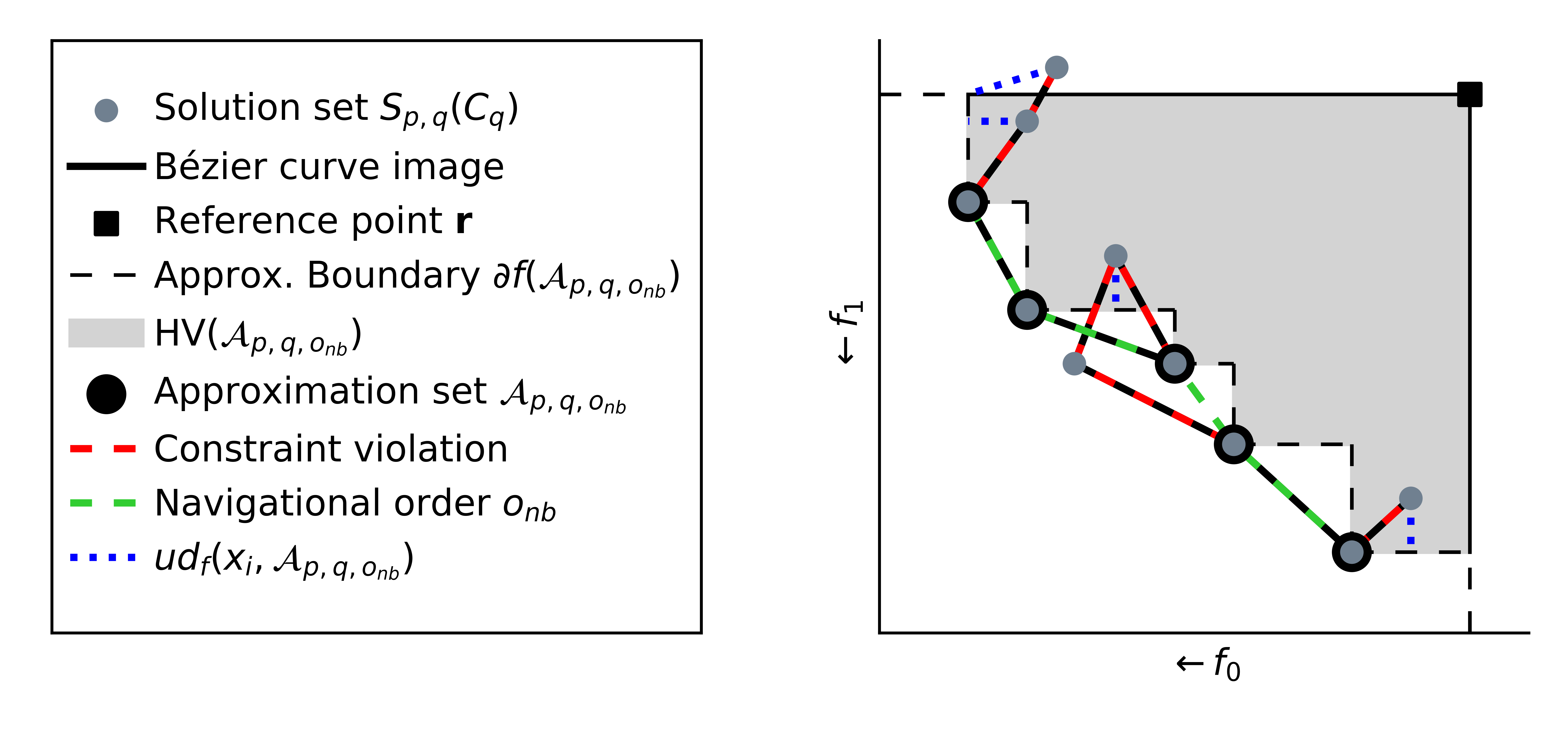}
  \caption{Evaluation of Bézier parameterizations \cite{bezea}}
  \label{fig:bezier_eval}
\end{wrapfigure}

To evaluate a solution set $S_{p,q}(C_q)$, a number of new functions were previously introduced \cite{bezea}.
These functions are briefly explained in the following paragraphs.
Figure \ref{fig:bezier_eval} illustrates these functions to give the reader a more graphical indication.

A new function $A^{nb}\left( S_{p,q} \right)$ has been introduced that calculates a navigational Bézier (nb) order $o_{nb}$.
This order is defined as starting from the best solution for objective $f_0$ to the best solution in $f_1$.
All solutions that are dominated by other solutions on the curve, are omitted from the subset that defines the navigational order.
An approximation set $\mathcal{A}_{p,q,o_{nb}}$ is the resulting subset of $S_{p,q}(C_q)$, with only the solution indices as specified in $o_{nb}$.
The quality of the approximation set $\mathcal{A}_{p,q,o_{nb}}$ can now be evaluated, e.g., with the HV indicator \cite{spea2}.

A new constraint function $C\left(\mathcal{S}_{p,q}, o_{nb}\right)$ was also introduced.
It is employed in order to not only push all dominated solutions on the curve towards the undominated region of the search space, but also to prevent the curve from intersecting itself in objective space. 
This may for instance happen if a curve stretches across two local PSs, which is not preferential.
The constraint function uses the uncrowded distance metric $ud_f(\boldsymbol{x}_i, \mathcal{A})$ \cite{uhvi}, which measures the Euclidean distance from a dominated point $\boldsymbol{x}_i$ to the approximation boundary $\partial f(\mathcal{A}_{p,q,o_{nb}})$ in objective space.
Furthermore, to further increase pressure towards the unfolding of Bézier curves in objective space, all dominated solutions and those not in $\mathcal{A}_{p,q,o_{nb}}$ are pulled towards their neighbouring solutions on the Bézier curve by taking the Euclidean distance in objective space between these solution and their neighbours as an additional constraint value.
All dominated solutions from $S_{p,q}(C_q)$ now have their uncrowded distance values and the Euclidean distances in objective space to neighbours of those not in $\mathcal{A}_{p,q,o_{nb}}$ summed up as a constraint for the total solution set.
In combination with constraint domination \cite{constraint_dom}, this constraint pushes all solutions along the Bézier curve towards the undominated region. 

\section{Niching methods}
To enable the algorithm proposed in this paper, i.e., MM-BezEA, to effectively search the multi-modal landscape, several previously introduced niching methods are used and combined.
These are employed in order to extend the uni-modal search that is originally performed by BezEA.
As the number of modes is usually unidentified beforehand, the algorithm needs to be able to adapt to the number of modes present in an MMOP.

\subsection{HVC and MO HVC}
HVC is a so-called two-stage niching approach that clusters and evolves the population for multi-modal single-objective optimization problems.
In each generation, the first stage is used to locate each of the distinct niches, for each of which a core search algorithm is initialized in the second stage.

At the heart of the HVC approach is the Hill-Valley Test (HVT) \cite{hv_test}, which can be utilized to determine whether two solutions reside in the same niche.
It first determines an edge between two solutions $\boldsymbol{x}_i$ and $\boldsymbol{x}_j$ in the search space.
Along this edge, $N_t$ evenly spread points are evaluated, determined by the distance between the two solutions divided by the expected edge length.
If any of these $N_t$ test points have a fitness that is worse than that of $\boldsymbol{x}_i$ and $\boldsymbol{x}_j$, the test detects that there is a \textit{hill} in between them.
Consequently, the two solutions are to be put in separate clusters.
On the other hand, if all $N_t$ points have equal or better fitness values than both $\boldsymbol{x}_i$ and $\boldsymbol{x}_j$, these two solutions belong to the same \textit{valley} and are to be clustered together.
In order to determine in which order the solutions are to be clustered (i.e., undergo the HVT), the concept of the nearest better tree \cite{nbt} is employed.

The MO-HillVallEA algorithm \cite{mohillvallea} expands on the previous HVC approach in the form of MO-HVC.
It uses the same concept of the HVT, but now performs clustering for each of the $m$ objective functions separately, which results in $m$ cluster sets.
To obtain a single cluster set, the intersection of each pair of clusters from all $m$ clustering sets is taken, similar to the colored regions of Figure \ref{fig:problem}.


\subsection{Restart scheme with elitist archive}
\label{sec:restart}
Various algorithms implemented a form of a restart scheme whereby the population size is increased over time.
Examples of such schemes are the interleaved multistart scheme \cite{mo-rv-gomea,param-free} and the restart-Covariance Matrix Adaptation Evolution Strategy with Increased Population (IPOP-CMA-ES) algorithm \cite{ipop-cma-es}.
In HillVallEA \cite{hillvallea}, an elitist archive is combined with a restart scheme, where the population size is doubled after each restart as in IPOP-CMA-ES \cite{ipop-cma-es}.
By employing the HVT to check if a solution resides in another niche, the elitist archive of HillVallEA is capable of holding the elites for each of the modes, despite it being developed for single objective problems.

To prevent HillVallEA from revisiting already searched modes, it makes use of the elitist archive, which is inspired by the repelling subpopulations (RS-CMSA) algorithm \cite{RS-CMSA} that defines taboo regions close to elites.
The steps taken to discard the regions of the search space, for which an elite was already found in one of the earlier populations, start with adding the elites to the population of the current restart.
Then, all solutions are clustered using HVC, followed by discarding all clusters that have one of the elites as their best solution.
As a result of discarding these regions of the search space, more attention is given to undiscovered parts of the search space after each restart.

\section{Multi Modal-Bézier Evolutionary Algorithm}
In this section MM-BezEA is described.
MM-BezEA is comprised of a combination and modification of techniques described in the previous sections.
The most notable of the modifications are the adjustments implemented in HVC in order to apply it to Bézier curve parameterizations, as well as the initialization of approximation sets within niches.

\subsection{Clustering approximation sets}
The Bézier curves are evaluated using the uncrowded HV measure \cite{uhvgomea}. 
Since this is a scalar, the HVC approach seems to intuitively allow the clustering of single-objective problems.
However, the approximation set $\mathcal{A}_{p,q,o_{nb}}$ that is used in the HV calculation only considers the undominated indices of the Bézier solution set $S_{p,q}(C_q)$ as defined in $o_{nb}$.
Hence, the objective value of a solution set $S_{p,q}(C_q)$ seems highly dependant on how many dominated solutions there are on the Bézier curve due to its orientation and length in decision space.

To enable the clustering of Bézier solution sets $S_{p,q}(C_q)$, the idea behind MO-HVC can be used on the set of control points $C_q$, as each of these is a single solution as normally defined in MO optimization.
Also, since a solution set is defined to be deteriorating in $f_0$ and improving in $f_1$ according to $o_{nb}$, the order of the control points is inverted if $f_0(\boldsymbol{c}_1) < f_0 (\boldsymbol{c}_q)$ does not hold \cite{bezea}.
Accordingly, the $i$-th Bézier solution can be designated to be in the same niche as the $j$-th Bézier solution if their control points $\boldsymbol{c}_l^i \in C_q^i$ and $\boldsymbol{c}_l^j \in C_q^j$ for $l = \left\{1,...,q\right\}$ are in the same niche.
In a general sense, the same HVC approach as used in HillVallEA is used, but inspiration has been taken from the MO-HVC approach to produce a new test for Bézier solution sets, which is shown in Algorithm \ref{alg:bez_hvt}.

\vspace{-7.5mm}
\begin{algorithm}
    \small
    \DontPrintSemicolon
    \Input{Solutions sets $\mathbf{S_i}$, $\mathbf{S_j}$, int $N_t$, objective functions $f_0,...,f_{m-1}$}
    \Output{Whether $\mathbf{S_i}$ and $\mathbf{S_j}$ belong to the same niche}
    \For{\texttt{$l = 1,..., q$}}{
        $\boldsymbol{c}_{i,l}, \boldsymbol{c}_{j,l} \gets $ control point $l$ of $\mathbf{S_i}$, control point $l$ of $\mathbf{S_j}$ \;
        \tcp*[l]{Check if $\boldsymbol{c}_{i,l}$ and $\boldsymbol{c}_{j,l}$ are in same niche for all $m$ objectives}
        \For{\texttt{$k = 0,...,m-1$}}{
            \lIf{HillValleyTest($\boldsymbol{c}_{i,l}$, $\boldsymbol{c}_{j,l}$, $N_t$, $f_k$)}{
                \Return false
            }
        }
    }
    \Return true
    \caption{[$B$] = Bezier-HillValleyTest($\mathbf{S_i}$, $\mathbf{S_j}$, $N_t$, $f$)}
    \label{alg:bez_hvt}
\end{algorithm}
\vspace{-10.5mm}

\subsection{Initialization within niches}
The original BezEA algorithm initializes all solution sets by sampling from a uniform distribution over the search space.
As it is an MOEA that was not designed for multi-modal optimization, the uniform initialization allows solution sets to be initialized within or in between any niche(s).
Clustering these solutions with the newly introduced Bézier HVT will result in finding a large number of separate niches as each control point has to be in the same mode.
To prevent this, a new initialization method for Bezier solution sets is proposed to enforce their initialization within a niche.
First, an iteration of MO-HVC
\begin{wrapfigure}{r}{0.45\textwidth}
  \vspace{-9.6mm}
  \centering
  \includegraphics[width=1.0\linewidth]{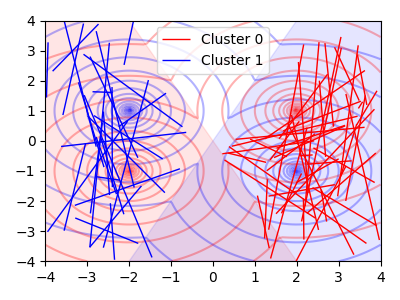}
  \setlength{\abovecaptionskip}{-5.5pt}
  \setlength{\belowcaptionskip}{-10pt}
  \caption{Initialization of Bézier solutions ($q=2$) for MinDist}
  \label{fig:bez_init}
  \vspace{-5mm}
\end{wrapfigure}
is run on a set of $q \times N$ solutions, $N$ being the population size, that is sampled from a uniform distribution over the search space, where the resulting clusters include all of the $\boldsymbol{x}_{test}$ solutions resulting from applying the Hill-Valley Test.
Second, selection is performed for each cluster proportional to their size in order to reduce their combined size, with test solutions, down to $q \times N$.
Lastly, Bézier solution sets $S_{p,q}(C_q)$ are initialized by randomly choosing $q$ solutions as the control points from one single cluster $\mathcal{C}$ as produced by MO-HVC if $|\mathcal{C}| \geq q$.
The result can be seen in Figure \ref{fig:bez_init} in the case of the example problem MinDist.

\subsection{Algorithm overview}
MM-BezEA has a similar structure as the restart scheme in HillVallEA \cite{hillvallea} that is described in section \ref{sec:restart}.
Every iteration, the combination of initialization of Bézier curves and dismissal of previously optimized clusters with an elite as their best solution takes place. 
For each of the resulting niches, a core search algorithm 
\begin{wrapfigure}[20]{r}{0.6\textwidth}
\vspace{-7mm}
\footnotesize
\setlength{\algomargin}{0pt}
\begin{algorithm}[H]
    \DontPrintSemicolon
    \Input{MO function $f$, popsize $N$, test points $p$, control points $q$, budget}
    \Output{Elitist archive $\mathbb{E} = [\mathcal{E}_0, \mathcal{E}_1, ...]$}
    $\mathbb{E} = \{\}$ \;
    \While{budget remaining}{ 
        $\mathcal{P}_{mo} =$ UniformSampling($q \times N$, $f$) \;
        $\mathbb{C}_{mo} =$ MO-HillValleyClustering($\mathcal{P}_{mo}$, $f$) \;
        $\mathbb{C} =$ InitializeBezierSolutions($\mathbb{C}_{mo}$, $q$, $p$, $f$) \;
        $\mathbb{C} =$ RemoveElitesFrom$\left( \mathbb{C} \right)$\;
        \While{budget remaining}{
            $\mathcal{P} = \mathbb{E}$ \;
            \For{$\mathcal{C}_i \in \mathbb{C}$}{
                $\mathcal{O}_i =$ core\_search\_algorithm$\left( \mathcal{C}_i \right)$ \;
                $\mathcal{P} = \mathcal{P} \cup \mathcal{O}_i$ \;
            }
            $\mathbb{C}_{prev} = \mathbb{C}$ \;
            $\mathbb{C} =$ BezierHillValleyClustering$\left( \mathcal{P}, f \right)$\;
            $\mathbb{E} =$ ConstructElitistArchive$\left( \mathbb{C}, \mathbb{E} \right)$ \;
            $\mathbb{C} =$ RemoveElitesFrom$\left( \mathbb{C} \right)$\;
            $\mathbb{C} =$ ClusterRegistration$\left( \mathbb{C}, \mathbb{C}_{prev} \right)$\;
        }
        }
    \caption{[$\mathbb{E}$] = MM-BezEA$(...)$} 
    \label{alg:mm_bezea}
\end{algorithm}
\end{wrapfigure}
is run for one generation, which in the case of MM-BezEA is the RV-GOMEA algorithm \cite{rv_gomea} that is also used in BezEA.
At the end of each generation, the Bézier HVT of Algorithm \ref{alg:bez_hvt} is used in the HVC step.
This step takes all solutions originating from all clusters and clusters them again for the next generation.
In between generations, the notion of cluster registration \cite{cluster_reg} is used on the cluster mean closest in decision space to transfer the parameters for RV-GOMEA between the clusters of each generation.
An overview of the algorithm in the form of pseudocode is given in Algorithm \ref{alg:mm_bezea}.
\section{Experiments}
MM-BezEA is empirically benchmarked on several test problems.
The results are compared to MO-HillVallEA \cite{mohillvallea}, MO-RV-GOMEA \cite{mo-rv-gomea}, and MO\_Ring\_PSO\_SCD \cite{mo-pso-ring}. 
MO\_Ring\-\_PSO\_SCD is implemented through the PlatEMO framework \cite{PlatEMO}, together with a manual implementation of the used metrics and problems.
For the other algorithms, original C++ implementations are used.

\subsection{Test problems}
Several test problems are employed. 
First of these is the MinDist problem \cite{mohillvallea} that was described in the introduction, where linear PSs are to be found.
The other employed functions are frequently used in literature, namely OmniTest \cite{omni-optimizer}, Two on One \cite{twoonone}, and Sympart \{1,2,3\} \cite{sympart}.
Lastly, several problems are 
\begin{wraptable}{r}{0.65\textwidth}
\vspace{-1.5mm}
\small
\centering
\caption{Bi-objective problem characteristics.}
\begin{tabular}{|l|l|l|l|l|}
\hline
Problem                    & $\ell$  & PS     & PS Shape   & PF Shape     \\ \Xhline{3\arrayrulewidth}
MinDist                    & $[2,\infty) \in \mathbb{Z}$ & n        & Linear     & Convex       \\ \hline
Omni Test                  & $[2,\infty) \in \mathbb{Z}$ & $3^\ell$ & Linear     & Convex       \\ \hline
Two on One                 & 2       & 2        & Linear     & Convex       \\ \hline
Sympart 1, 2             & 2       & 9        & Linear     & Convex       \\ \hline
Sympart 3                  & 2       & 9        & Non-linear & Convex       \\ \hline
MMF 1, 2                 & 2       & 2        & Non-linear & Convex       \\ \hline
MMF 12                     & $[2,\infty) \in \mathbb{Z}$ & n        & Linear     & Disconnected \\ \hline
MMF 14, 15               & $[2,\infty) \in \mathbb{Z}$ & n        & Linear     & Concave      \\ \hline
\end{tabular}
\vspace{-7.5mm}
\label{tab:problems}
\end{wraptable}
taken from the Multi-modal Multi-objective test Function (MMF) benchmark suite \cite{mmf} in the form of MMF \{1, 2, 12, 14, 15\}.
A mix of PS and PF shapes have been chosen to determine the capabilities of MM-BezEA on different problem types.
Table \ref{tab:problems} shows some of the important characteristics for each of the problems.
For all problems with a configurable number of PSs $n$, it is set to 2, likewise the problem dimension $\ell$ is fixed to 2.
In order to determine the values of the performance indicators, the reference PSs will be made using 5000 points that adhere to the analytical formulas describing the PSs.
In the case of Two on One, a very close approximation is used \cite{twoonone}.

\subsection{Benchmark setup}
In order to get a fair comparison, each of the algorithms will be given an equally sized budget of $200,000$ function evaluations for each of the problems.
This removes the influence of the used programming languages, as the computation time is not limited.
The parameters of MO-RV-GOMEA, MO\_Ring\_PSO\_SCD, and MO-HillVallEA are set to the values reported in relevant literature.
Furthermore, for each problem and metric, the average over 31 runs will be taken.

The elitist archives sizes $N_\mathbb{E}$ are set to be 1250 for MO-RV-GOMEA and MO-HillVallEA.
The population size $N$ is set to 96 for MO-RV-GOMEA and 250 for MO-HillVallEA \cite{mohillvallea}.
MO-RV-GOMEA uses a linkage tree as its linkage model, with a total of 5 clusters.
For MO\_Ring\_PSO\_SCD the population size is 800 \cite{mo-pso-ring}. 
For the MM-BezEA algorithm, the number of control points $q$ for each approximation set is set to 2.
Just like for the original BezEA algorithm, MM-BezEA is given population sizes of 76 \cite{bezea}.
The number of test points $p$ is set to 7.

\subsection{Performance indicators}
The HV indicator \cite{spea2} is used to see how well the algorithms perform in getting close the PF.
As a result of the use of test points in MM-BezEA, the Bézier solutions sets will have a limited amount of points in the approximation set that can be used to calculate the HV values.
Therefore, a subset of the approximation set will be taken for the other algorithms to allow a fair comparison based on the HV indicator.
Specifically, the same number of test points is selected for a fair comparison by means of greedy Hypervolume Subset Selection (gHSS) \cite{ghss}.

We further use a relatively new performance indicator for multi-modal multi objective optimization, named Pareto Set Proximity (PSP) \cite{mo-pso-ring}.
It is an indicator that determines how well all PSs are approximated by taking the Cover Rate (CR), that shows how well the extremes of all PSs are captured, divided by the Inverted Generational Distance in decision space (IGDX) \cite{igdx}, which can be used to determine how close the approximation sets are to the PSs.
For the IGDX measure, the approximation sets as produced by MM-BezEA are interpolated by taking 1000 intermediate points before determining the IGDX value.
This can be performed relatively easily as interpolating these parameterizations does not require any extra fitness evaluations.

Finally, we use a performance indicator regarding smoothness, for which we follow the definition as introduced in the work on BezEA \cite{bezea}. 
It captures how smooth an approximation set can be navigated in terms of decision variables by measuring the detour length in decision space when traversing the approximation set from one solution to the next via an intermediate solution, as compared to going to the next solution directly.
The smoothness approaches its maximum value of 1 if all solutions would be colinear in decision space, where the lowest possible value is 0.
In cases where multiple approximation sets are explicitely determined, like in MO-HillVallEA and MM-BezEA, the average smoothness over all clusters will be taken.
In the other cases the smoothness over the entire approximation set is taken.


\subsection{Results}
Table \ref{tab:res_hv} shows the results for all problems and algorithms per indicator.

\begin{table*}[t]
\renewcommand{\bfdefault}{b}
\centering
\caption{Results (avg. ($\pm$ st.dev.)) per problem and algorithm over 31 runs, bold identifies best result with statistical significance (Wilcoxon rank-sum test with $\alpha = 0.05$ and Holm-Bonferroni correction).}
\label{tab:res_hv}
\scriptsize
\begin{tabular}{l|l|lllll}
\hline
  & Problem   & MM-BezEA                      & MO-HillVallEA                 & MO\_Ring\_PSO\_SCD               & MO-RV-GOMEA                   \\
\hline
 \multirow{11}{*}{\rotatebox[origin=c]{90}{~\large{HV}}} & MinDist   & \textbf{1.17e+2 (±9.43e-5)} & 1.17e+2 (±1.62e-2)          & 1.17e+2 (±5.95e-3)          & 1.17e+2 (±1.36e-3)           \\
 & OmniTest  & \textbf{8.48e+0 (±2.18e-6)} & 8.47e+0 (±1.96e-3)          & 8.47e+0 (±7.34e-4)          & 8.47e+0 (±4.82e-4)          \\
 & Sympart 1  & \textbf{1.17e+2 (±2.16e-5)} & 1.17e+2 (±1.42e-2)          & 1.17e+2 (±7.64e-3)          & 1.17e+2 (±1.30e-3)          \\
 & Sympart 2  & \textbf{1.17e+2 (±2.91e-5)} & 1.17e+2 (±7.77e-3)          & 1.17e+2 (±8.75e-3)          & 1.17e+2 (±4.47e-3)          \\
 & Sympart 3  & \textbf{1.17e+2 (±9.61e-5)} & 1.17e+2 (±1.61e-2)          & 1.17e+2 (±9.21e-3)          & 1.17e+2 (±4.91e-3)          \\
 & TwoOnOne  & 1.13e+2 (±1.33e-4)          & 1.13e+2 (±2.39e-4)          & \textbf{1.13e+2 (±1.82e-4)} & 1.13e+2 (±1.10e-4)          \\
 & MMF 1     & 6.04e-1 (±3.86e-2)          & \textbf{8.05e-1 (±2.37e-4)} & \textbf{8.05e-1 (±8.69e-5)} & \textbf{8.05e-1 (±6.60e-5)} \\
 & MMF 2     & 6.34e-1 (±2.01e-4)          & 8.04e-1 (±6.90e-4)          & 8.04e-1 (±9.59e-4)          & \textbf{8.05e-1 (±1.75e-4)} \\
 & MMF 12     & 1.78e+0 (±2.02e-6)          & 2.06e+0 (±2.57e-3)          & 2.06e+0 (±2.05e-3)          & \textbf{2.06e+0 (±1.49e-4)} \\
 & MMF 14     & 5.63e+0 (±1.33e-5)          & \textbf{5.63e+0 (±7.26e-4)} & 5.63e+0 (±1.92e-3)          & 5.63e+0 (±2.23e-4)          \\
 & MMF 15     & 5.56e+0 (±2.03e-2)          & \textbf{5.57e+0 (±6.52e-4)} & 5.56e+0 (±1.54e-3)          & \textbf{5.57e+0 (±1.79e-4)} \\
\hline
  \multirow{11}{*}{\rotatebox[origin=c]{90}{~\large{PSP}}} &MinDist   & \textbf{3.26e+2 (±7.31e+1)} & 5.02e+1 (±2.74e+0)          & 6.97e+1 (±6.73e+0)          & 1.21e+0 (±1.65e+0) \\
  & OmniTest  & \textbf{2.36e+2 (±8.81e+1)} & 7.13e+1 (±2.76e+0)          & 6.90e+1 (±9.56e+0)          & 1.36e-1 (±2.43e-1) \\
 &Sympart 1  & \textbf{2.67e+2 (±1.21e+2)} & 3.56e+1 (±1.57e+0)          & 2.79e+1 (±3.69e+0)          & 1.16e-2 (±2.70e-2) \\
 &Sympart 2  & \textbf{3.09e+2 (±8.46e+1)} & 3.60e+1 (±7.42e-1)          & 2.38e+1 (±2.46e+0)          & 1.02e-2 (±1.71e-2) \\
 &Sympart 3  & 6.41e+1 (±7.77e+1)          & \textbf{4.33e+1 (±2.10e+0)} & 2.64e+1 (±5.62e+0)          & 8.09e-3 (±1.33e-2) \\
 &TwoOnOne  & \textbf{3.04e+2 (±2.18e+2)} & 4.50e+1 (±7.21e-1)          & 2.45e+1 (±1.03e+1)          & 2.68e+0 (±1.11e+0) \\
 &MMF 1     & 7.22e+0 (±2.41e+0)          & 3.17e+1 (±6.84e-1)          & \textbf{3.80e+1 (±6.79e+0)} & 1.02e+0 (±2.89e-1) \\
 &MMF 2     & 4.00e+0 (±2.49e+0)          & \textbf{1.17e+2 (±1.21e+1)} & 5.04e+1 (±1.51e+1)          & 2.18e+0 (±1.41e+0) \\
 &MMF 12     & \textbf{2.42e+1 (±7.72e+0)} & 1.94e+1 (±6.46e+0)          & 1.53e+1 (±1.46e-1)          & 8.67e+0 (±1.36e+0) \\
 &MMF 14     & \textbf{2.68e+3 (±4.82e+2)} & 3.70e+2 (±1.03e+1)          & 2.31e+2 (±2.16e+1)          & 1.08e+0 (±2.84e+0) \\
 &MMF 15     & \textbf{2.73e+2 (±4.17e+1)} & 2.65e+2 (±4.58e+0)          & 2.44e+2 (±7.25e+0)          & 2.24e+1 (±1.40e-2) \\
\hline
  \multirow{11}{*}{\rotatebox[origin=c]{90}{~\large{Smoothness}}} & MinDist   & \textbf{1.00e+0 (±0.00e+0)} & 8.09e-1 (±3.70e-2) & 7.63e-1 (±5.91e-3) & 8.94e-1 (±1.81e-1)          \\
  & OmniTest   & \textbf{1.00e+0 (±0.00e+0)} & 9.23e-1 (±5.65e-3) & 7.28e-1 (±9.76e-3) & 7.16e-1 (±1.96e-1)           \\
 & Sympart 1  & \textbf{1.00e+0 (±0.00e+0)} & 8.76e-1 (±2.52e-2) & 6.84e-1 (±9.76e-3) & 7.01e-1 (±2.09e-1)          \\
 & Sympart 2  & \textbf{1.00e+0 (±0.00e+0)} & 8.78e-1 (±2.14e-2) & 5.73e-1 (±9.30e-3) & 7.87e-1 (±1.68e-1)          \\
 & Sympart 3  & \textbf{1.00e+0 (±0.00e+0)} & 8.70e-1 (±3.10e-2) & 5.29e-1 (±1.18e-2) & 8.30e-1 (±1.92e-1)          \\
 & TwoOnOne   & \textbf{1.00e+0 (±0.00e+0)} & 7.77e-1 (±1.21e-2) & 7.47e-1 (±9.36e-3) & 7.51e-1 (±1.59e-1)          \\
 & MMF 1      & \textbf{1.00e+0 (±0.00e+0)} & 9.01e-1 (±3.98e-2) & 7.82e-1 (±8.49e-3) & 5.86e-1 (±9.13e-2)          \\
 & MMF 2      & \textbf{1.00e+0 (±0.00e+0)} & 9.38e-1 (±1.26e-2) & 5.01e-1 (±8.92e-3) & 8.46e-1 (±1.60e-1)          \\
 & MMF 12     & \textbf{1.00e+0 (±0.00e+0)} & 8.35e-1 (±2.73e-2) & 6.35e-1 (±9.86e-3) & \textbf{1.00e+0 (±0.00e+0)} \\
 & MMF 14     & \textbf{1.00e+0 (±0.00e+0)} & 9.31e-1 (±1.17e-2) & 8.71e-1 (±1.11e-2) & 9.62e-1 (±1.05e-1)          \\
 & MMF 15     & \textbf{1.00e+0 (±0.00e+0)} & 9.19e-1 (±1.43e-2) & 8.63e-1 (±8.54e-3) & \textbf{1.00e+0 (±0.00e+0)} \\
\hline
\end{tabular}
\end{table*}

The HV results clearly show that all algorithms are capable of performing nearly equally in obtaining a good approximation front. 
However, MM-BezEA with $q = 2$ does deteriorate in performance on the MMF1 and 2 problems that have non-linear PSs.
The deterioration is inherently caused by the chosen parameterizations that create approximation sets which are linear in shape.
Another problem instance where a smaller HV for the new algorithm is obtained, is that of MMF12.
Here, despite MM-BezEA obtaining the best PSP values, the approximation sets did not fully approximate the actual PSs and did not cover the endpoints.

\begin{figure}
    \centering
     \begin{subfigure}[t]{0.49\textwidth}
         \centering
         \includegraphics[width=\linewidth]{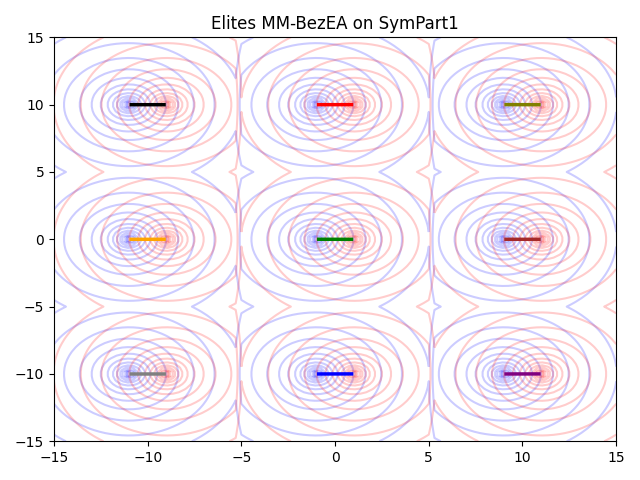}
         \caption{All approximation sets produced by MM-BezEA on SymPart 1 \cite{sympart}}
         \label{fig:elites}
     \end{subfigure}
     \hfill
     \begin{subfigure}[t]{0.49\textwidth}
         \centering
         \includegraphics[width=\linewidth]{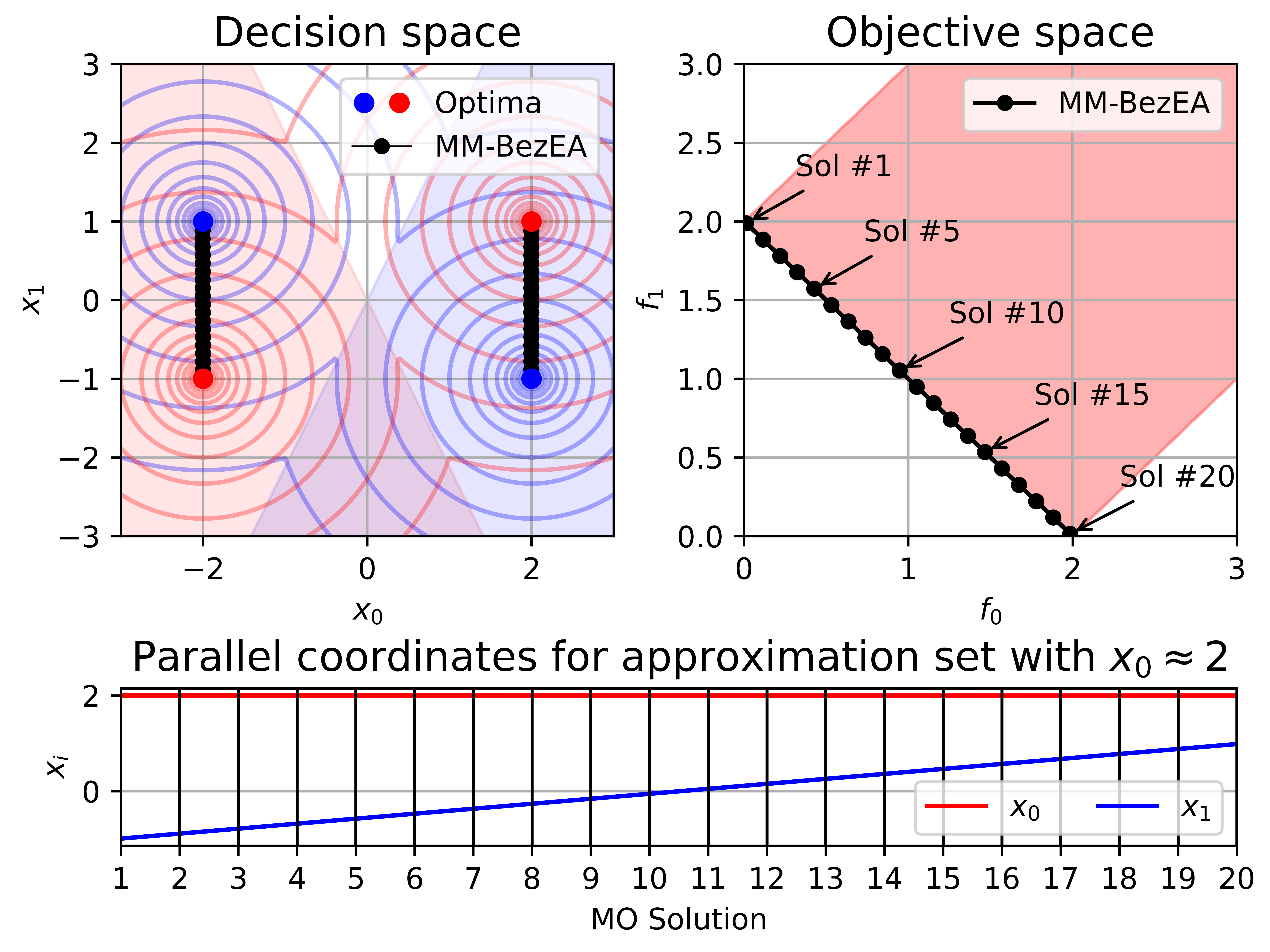}
         \caption{Approximation sets and front with parallel coordinates plot for one of the approximation sets produced by MM-BezEA on MinDist \cite{mohillvallea}}
         \label{fig:smooth_results}
     \end{subfigure}
     \caption{Visualization of results}
\end{figure}

The PSP indicator shows similar results, except that MO-RV-GOMEA performs worse as it is not an MMOEA and therefore does not explicitly search for multiple niches.
Again promising results for MM-BezEA are shown in cases where linear PSs can be found, as seen in Figure \ref{fig:elites} where MM-BezEA approximates all 9 Pareto sets very well.
In the problems with non-linear PSs, MO-HillVallEA and MO\_Ring\_PSO\_SCD manage to find better approximations.

The smoothness results show, as intended, that the chosen parameterizations inherently cause smooth approximation sets with a perfect smoothness of 1.0 for MM-BezEA.
Other algorithms do not obtain this, except for MO-RV-GOMEA on 2 of the 11 problems. A visualization of the results of MM-BezEA on the MinDist problem is given in Figure \ref{fig:smooth_results}.
This figure depicts the smooth progression of the decision variables values in the parallel coordinates plot for the rightmost approximation set in decision space.
It contrasts sharply to the parallel coordinates plot of Figure \ref{fig:problem} when navigating the approximation set as it now shows a smooth course of the decision variable values.

\section{Discussion}
MM-BezEA did not cover the endpoints of the PSs in the case of MMF12. 
This can be caused by the fact that the Bézier fitness function will constrain a solution when one of its control points is dominated in objective space by one of the test points.
As the endpoints of each part of the discontinuous PF are close to being dominated, i.e., close to the constraint space, it can lead to not entirely capturing the discontinuous pieces of the PF and thus resulting in a lower HV.

The HV indicator is a Pareto compliant indicator \cite{hv_par,hv_ind_comp}, but it does suffer from a downside.
In some situations the endpoints of the approximation sets cannot reach the endpoints of the Pareto set because the distribution of points that maximizes the hypervolume does not include the extreme solutions.
Even when the number of test points will be set to infinity, the reference point can never be set so that the extremes are captured \cite{ref_point}.

Even though the smoothness indicator tries to determine whether an approximation set is smooth by measuring the detour length, it comes down to determining the angle between neighboring solutions.
This implies that it only considers linear curves to be perfectly smooth, where there is a straight angle between solutions.
When the number of solutions in an approximation set increases, the average distance between the solutions in objective space decreases.
As a result of the lower distances and ever so slight oscillations around the PF, the angle between solutions decreases due to which the smoothness indicator will report low smoothness values.
In cases where the niches can be separated in a good manner, another definition of smoothness that considers the oscillation around the PS might be more useful.

Future work could investigate the further use of the Bézier parameterizations with more control points to approximate non-linear Pareto sets.
Note that the definitions given in this paper already allow for this.
Furthermore, no limit on the number of approximation sets can currently be set, which degrades the quality of the approximation sets in highly multi-modal problems as the population is then divided over all niches through HVC \cite{mohillvallea}.
Finally, Bézier simplexes \cite{beziersimplex} might be usable for problems with more than two objectives.
\vspace{-2mm}
\section{Conclusion}
We proposed the algorithm MM-BezEA to search for multiple parameterized approximation sets that define smooth curves in the decision space for bi-objective multi-modal optimization problems.
The results show that MM-BezEA is competently capable of locating all modes in a multi-modal landscape as exemplified in various benchmark problems and that the smoothness is indeed enforced by the Bézier parameterizations.
Furthermore, MM-BezEA significantly outperformed other algorithms in problems with linear Pareto sets, but was outperformed in problems with non-linear Pareto sets.
However, only low-order Bézier curves were used in our experiments, and these results may well be different if higher order curves were used, which the definitions in this paper readily allow.
\bibliographystyle{splncs04}
\bibliography{bibliography}

\begin{thebibliography}{10}
\providecommand{\url}[1]{\texttt{#1}}
\providecommand{\urlprefix}{URL }
\providecommand{\doi}[1]{https://doi.org/#1}

\bibitem{RS-CMSA}
Ahrari, A., Deb, K., Preuss, M.: Multimodal optimization by covariance matrix
  self-adaptation evolution strategy with repelling subpopulations.
  Evolutionary Computation  \textbf{25}(3),  439–471 (Sep 2017).
  \doi{10.1162/evco\_a\_00182}

\bibitem{ref_point}
Auger, A., Bader, J., Brockhoff, D., Zitzler, E.: Theory of the hypervolume
  indicator: Optimal $\mu$-distributions and the choice of the reference point.
  In: Proceedings of the tenth ACM SIGEVO workshop on Foundations of genertic
  algorithms. p. 87–102. FOGA '09, Association for Computing Machinery, New
  York, NY, USA (2009). \doi{10.1145/1527125.1527138}

\bibitem{ipop-cma-es}
Auger, A., Hansen, N.: A restart {CMA} evolution strategy with increasing
  population size. In: 2005 IEEE Congress on Evolutionary Computation. vol.~2,
  pp. 1769--1776 Vol. 2. IEEE, New York, NY, USA (2005).
  \doi{10.1109/CEC.2005.1554902}

\bibitem{hv_vs_pareto}
Berghammer, R., Friedrich, T., Neumann, F.: Convergence of set-based
  multi-objective optimization, indicators and deteriorative cycles.
  Theoretical Computer Science  \textbf{456},  2--17 (Oct 2012).
  \doi{10.1016/J.TCS.2012.05.036}

\bibitem{smsemoa}
Beume, N., Naujoks, B., Emmerich, M.: {SMS-EMOA}: Multiobjective selection
  based on dominated hypervolume. European Journal of Operational Research
  \textbf{181}(3),  1653--1669 (Sep 2007).
  \doi{https://doi.org/10.1016/j.ejor.2006.08.008}

\bibitem{cluster_reg}
Bosman, P.A.N.: The anticipated mean shift and cluster registration in
  mixture-based {EDAs} for multi-objective optimization. In: Proceedings of the
  12th Annual Conference on Genetic and Evolutionary Computation. p. 351–358.
  GECCO '10, Association for Computing Machinery, New York, NY, USA (2010).
  \doi{10.1145/1830483.1830549}

\bibitem{rv_gomea}
Bouter, A., Alderliesten, T., Witteveen, C., Bosman, P.A.N.: Exploiting linkage
  information in real-valued optimization with the real-valued gene-pool
  optimal mixing evolutionary algorithm. In: Proceedings of the Genetic and
  Evolutionary Computation Conference. p. 705–712. GECCO '17, Association for
  Computing Machinery, New York, NY, USA (2017). \doi{10.1145/3071178.3071272}

\bibitem{mo-rv-gomea}
Bouter, A., Luong, N.H., Witteveen, C., Alderliesten, T., Bosman, P.A.N.: The
  multi-objective real-valued gene-pool optimal mixing evolutionary algorithm.
  In: Proceedings of the Genetic and Evolutionary Computation Conference. p.
  537–544. GECCO '17, Association for Computing Machinery, New York, NY, USA
  (2017). \doi{10.1145/3071178.3071274}

\bibitem{nsga2}
Deb, K., Pratap, A., Agarwal, S., Meyarivan, T.: A fast and elitist
  multiobjective genetic algorithm: {NSGA-II}. IEEE Transactions on
  Evolutionary Computation  \textbf{6},  182--197 (4 2002).
  \doi{10.1109/4235.996017}

\bibitem{constraint_dom}
Deb, K.: An efficient constraint handling method for genetic algorithms.
  Computer Methods in Applied Mechanics and Engineering  \textbf{186}(2),
  311--338 (Jun 2000). \doi{10.1016/S0045-7825(99)00389-8}

\bibitem{deb_moea}
Deb, K.: Multi-Objective Optimization Using Evolutionary Algorithms. John Wiley
  \& Sons, Inc., USA (2001)

\bibitem{omni-optimizer}
Deb, K., Tiwari, S.: Omni-optimizer: A generic evolutionary algorithm for
  single and multi-objective optimization. European Journal of Operational
  Research  \textbf{185}(3),  1062--1087 (Mar 2008).
  \doi{https://doi.org/10.1016/j.ejor.2006.06.042}

\bibitem{hv_par}
Fleischer, M.: The measure of {P}areto optima applications to multi-objective
  metaheuristics. In: Proceedings of the 2nd International Conference on
  Evolutionary Multi-Criterion Optimization. p. 519–533. EMO'03,
  Springer-Verlag, Berlin, Heidelberg (Apr 2003)

\bibitem{mmop}
Grimme, C., Kerschke, P., Aspar, P., Trautmann, H., Preuss, M., Deutz, A.H.,
  Wang, H., Emmerich, M.: Peeking beyond peaks: {C}hallenges and research
  potentials of continuous multimodal multi-objective optimization. Computers
  \& Operations Research  \textbf{136},  105489 (2021).
  \doi{https://doi.org/10.1016/j.cor.2021.105489}

\bibitem{ghss}
Guerreiro, A.P., Fonseca, C.M., Paquete, L.: Greedy hypervolume subset
  selection in low dimensions. Evolutionary Computation  \textbf{24},  521--544
  (Sep 2016). \doi{10.1162/EVCO\_a\_00188}

\bibitem{param-free}
Harik, G.R., Lobo, F.G.: A parameter-less genetic algorithm. In: Proceedings of
  the 1st Annual Conference on Genetic and Evolutionary Computation - Volume 1.
  p. 258–265. GECCO'99, Morgan Kaufmann Publishers Inc., San Francisco, CA,
  USA (1999)

\bibitem{MO-CMA-ES}
Igel, C., Hansen, N., Roth, S.: Covariance matrix adaptation for
  multi-objective optimization. Evolutionary Computation  \textbf{15},  1--28
  (3 2007). \doi{10.1162/evco.2007.15.1.1}

\bibitem{beziersimplex}
Kobayashi, K., Hamada, N., Sannai, A., Tanaka, A., Bannai, K., Sugiyama, M.:
  Bézier simplex fitting: Describing {P}areto fronts of simplicial problems
  with small samples in multi-objective optimization. In: Proceedings of the
  33rd AAAI Conference on Artificial Intelligence, AAAI 2019, the 31st
  Innovative Applications of Artificial Intelligence Conference, IAAI 2019 and
  the 9th AAAI Symposium on Educational Advances in Artificial Intelligence,
  EAAI 2019. pp. 2304--2313. AAAI press, Palo Alto, CA, USA (Jan 2019)

\bibitem{niching}
Li, X., Epitropakis, M.G., Deb, K., Engelbrecht, A.: Seeking multiple
  solutions: An updated survey on niching methods and their applications. IEEE
  Transactions on Evolutionary Computation  \textbf{21}(4),  518--538 (Aug
  2017). \doi{10.1109/TEVC.2016.2638437}

\bibitem{mmf}
Liang, J., Qu, B., Gong, D., Yue, C.: Problem definitions and evaluation
  criteria for the cec 2019 special session on multimodal multiobjective
  optimization. Tech. rep. (Nov 2018)

\bibitem{Luong2018}
Luong, N.H., Alderliesten, T., Bel, A., Niatsetski, Y., Bosman, P.A.N.:
  Application and benchmarking of multi-objective evolutionary algorithms on
  high-dose-rate brachytherapy planning for prostate cancer treatment. Swarm
  and Evolutionary Computation  \textbf{40},  37--52 (6 2018).
  \doi{10.1016/j.swevo.2017.12.003}

\bibitem{Mahfoud95nichingmethods}
Mahfoud, S.W.: Niching Methods for Genetic Algorithms. Ph.D. thesis, University
  of Illinois at Urbana-Champaign, USA (1996), uMI Order No. GAX95-43663

\bibitem{mohillvallea}
Maree, S.C., Alderliesten, T., Bosman, P.A.N.: Real-valued evolutionary
  multi-modal multi-objective optimization by hill-valley clustering. In:
  Proceedings of the Genetic and Evolutionary Computation Conference. p.
  568–576. GECCO '19, Association for Computing Machinery, New York, NY, USA
  (2019). \doi{10.1145/3321707.3321759}

\bibitem{bezea}
Maree, S.C., Alderliesten, T., Bosman, P.A.N.: Ensuring smoothly navigable
  approximation sets by {B}{\'e}zier curve parameterizations in evolutionary
  bi-objective optimization. In: Parallel Problem Solving from Nature -- PPSN
  XVI. pp. 215--228. Springer International Publishing, Cham (2020)

\bibitem{uhvgomea}
Maree, S.C., Alderliesten, T., Bosman, P.A.N.: {Uncrowded Hypervolume-based
  Multi-objective Optimization with Gene-pool Optimal Mixing}. Evolutionary
  Computation pp. 1--24 (Dec 2021). \doi{10.1162/evco\_a\_00303}

\bibitem{hillvallea}
Maree, S.C., Alderliesten, T., Thierens, D., Bosman, P.A.N.: Real-valued
  evolutionary multi-modal optimization driven by hill-valley clustering. In:
  Proceedings of the Genetic and Evolutionary Computation Conference. p.
  857–864. GECCO '18, Association for Computing Machinery, New York, NY, USA
  (Jul 2018). \doi{10.1145/3205455.3205477}

\bibitem{nbt}
Preuss, M.: Niching the {CMA-ES} via nearest-better clustering. In: Proceedings
  of the 12th Annual Conference Companion on Genetic and Evolutionary
  Computation. p. 1711–1718. GECCO '10, Association for Computing Machinery,
  New York, NY, USA (2010). \doi{10.1145/1830761.1830793}

\bibitem{twoonone}
Preuss, M., Naujoks, B., Rudolph, G.: Pareto set and {EMOA} behavior for simple
  multimodal multiobjective functions. In: Parallel Problem Solving from Nature
  -- PPSN IX. p. 513–522. PPSN'06, Springer-Verlag, Berlin, Heidelberg
  (2006). \doi{10.1007/11844297\_52}

\bibitem{sympart}
Rudolph, G., Naujoks, B., Preuss, M.: Capabilities of {EMOA} to detect and
  preserve equivalent pareto subsets. In: Evolutionary Multi-Criterion
  Optimization, 4th International Conference, {EMO} 2007, Matsushima, Japan,
  March 5-8, 2007, Proceedings. Lecture Notes in Computer Science, vol.~4403,
  pp. 36--50. Springer, Berlin, Heidelberg (2006).
  \doi{10.1007/978-3-540-70928-2\_7}

\bibitem{mmop_viz}
Sch{\"a}permeier, L., Grimme, C., Kerschke, P.: To boldly show what no one has
  seen before: A dashboard for visualizing multi-objective landscapes. In:
  Ishibuchi, H., Zhang, Q., Cheng, R., Li, K., Li, H., Wang, H., Zhou, A.
  (eds.) Evolutionary Multi-Criterion Optimization. pp. 632--644. Springer
  International Publishing, Cham (2021)

\bibitem{tanabe_review}
Tanabe, R., Ishibuchi, H.: A review of evolutionary multimodal multiobjective
  optimization. IEEE Transactions on Evolutionary Computation  \textbf{24}(1),
  193--200 (2020). \doi{10.1109/TEVC.2019.2909744}

\bibitem{PlatEMO}
Tian, Y., Cheng, R., Zhang, X., Jin, Y.: {PlatEMO}: A {MATLAB} platform for
  evolutionary multi-objective optimization. IEEE Computational Intelligence
  Magazine  \textbf{12}(4),  73--87 (Nov 2017)

\bibitem{uhvi}
Tour\'{e}, C., Hansen, N., Auger, A., Brockhoff, D.: Uncrowded hypervolume
  improvement: {COMO-CMA-ES} and the sofomore framework. In: Proceedings of the
  Genetic and Evolutionary Computation Conference. p. 638–646. GECCO '19,
  Association for Computing Machinery, New York, NY, USA (2019).
  \doi{10.1145/3321707.3321852}

\bibitem{hv_test}
Ursem, R.: Multinational evolutionary algorithms. In: Proceedings of the 1999
  Congress on Evolutionary Computation-CEC99. vol.~3, pp. 1633--1640 Vol. 3.
  IEEE, New York, NY, USA (1999). \doi{10.1109/CEC.1999.785470}

\bibitem{mo-pso-ring}
Yue, C., Qu, B., Liang, J.: A multi-objective particle swarm optimizer using
  ring topology for solving multimodal multi-objective problems. IEEE
  Transactions on Evolutionary Computation  \textbf{22},  805--817 (Sep 2017).
  \doi{10.1109/TEVC.2017.2754271}

\bibitem{igdx}
Zhang, Q., Li, H.: {MOEA/D}: A multiobjective evolutionary algorithm based on
  decomposition. IEEE Transactions on Evolutionary Computation  \textbf{11},
  712--731 (12 2007). \doi{10.1109/TEVC.2007.892759}

\bibitem{spea2}
Zitzler, E., Laumanns, M., Thiele, L.: {SPEA2}: Improving the strength pareto
  evolutionary algorithm for multiobjective optimization. In: Evolutionary
  Methods for Design Optimization and Control with Applications to Industrial
  Problems. pp. 95--100. International Center for Numerical Methods in
  Engineering, Athens, Greece (Sep 2001)

\bibitem{hv_ind_comp}
Zitzler, E., Thiele, L., Laumanns, M., Fonseca, C.M., da~Fonseca, V.G.:
  Performance assessment of multiobjective optimizers: an analysis and review.
  IEEE Transactions on Evolutionary Computation  \textbf{7}(2),  117--132 (May
  2003). \doi{10.1109/TEVC.2003.810758}

\end{thebibliography}


\begin{thebibliography}{8}
\bibitem{ref_article1}
Author, F.: Article title. Journal \textbf{2}(5), 99--110 (2016)

\bibitem{ref_lncs1}
Author, F., Author, S.: Title of a proceedings paper. In: Editor,
F., Editor, S. (eds.) CONFERENCE 2016, LNCS, vol. 9999, pp. 1--13.
Springer, Heidelberg (2016). \doi{10.10007/1234567890}

\bibitem{ref_book1}
Author, F., Author, S., Author, T.: Book title. 2nd edn. Publisher,
Location (1999)

\bibitem{ref_proc1}
Author, A.-B.: Contribution title. In: 9th International Proceedings
on Proceedings, pp. 1--2. Publisher, Location (2010)

\bibitem{ref_url1}
LNCS Homepage, \url{http://www.springer.com/lncs}. Last accessed 4
Oct 2017
\end{thebibliography}
\end{document}